\documentclass[letterpaper, 10 pt, conference]{ieeeconf}  
\IEEEoverridecommandlockouts                          
\usepackage{cite}
\usepackage{amsmath,amssymb,amsfonts}
\usepackage{algorithmic}
\usepackage{dsfont}
\usepackage{graphicx}
\usepackage{textcomp}
\usepackage{float}
\usepackage{multirow}
\usepackage[table,xcdraw]{xcolor}
\usepackage{todonotes}
\usepackage{hyperref}

\def\BibTeX{{\rm B\kern-.05em{\sc i\kern-.025em b}\kern-.08em
    T\kern-.1667em\lower.7ex\hbox{E}\kern-.125emX}}
\title{\LARGE \bf
PlaneSDF-based Change Detection for Long-term Dense Mapping}
\author{Jiahui Fu, Chengyuan Lin, Yuichi Taguchi, Andrea Cohen, Yifu Zhang, Stephen Mylabathula, John J. Leonard
\thanks{Jiahui Fu  and John Leonard are with the MIT Computer Science and Artificial Intelligence Laboratory, Cambridge, MA 02139, USA.
        {\tt\{jiahuifu,jleonard\}@mit.edu}}
\thanks{Chengyuan Lin, Yuichi Taguchi, Andrea Cohen, Yifu Zhang, and Stephen Mylabathula are with Meta, Menlo Park, CA 94025, USA. {\tt\{chengyuanlin, yuichitaguchi, Andy.Cohen, yifuzhang, mylabathula\}@fb.com}}        
\thanks{This work was done when Jiahui Fu was an intern at Meta.}}      

\begin{document}

\maketitle
\pagestyle{empty}
\thispagestyle{empty}

\begin{abstract}
The ability to process environment maps across multiple sessions is critical for robots operating over extended periods of time. Specifically, it is desirable for autonomous agents to detect changes amongst maps of different sessions so as to gain a conflict-free understanding of the current environment. In this paper, we look into the problem of change detection based on a novel map representation, dubbed Plane Signed Distance Fields (PlaneSDF), where dense maps are represented as a collection of planes and their associated geometric components in SDF volumes. Given point clouds of the source and target scenes, we propose a three-step PlaneSDF-based change detection approach:
(1) PlaneSDF volumes are instantiated within each scene and registered across scenes using plane poses; 2D height maps and object maps are extracted per volume via height projection and connected component analysis.
(2) Height maps are compared and intersected with the object map to produce a 2D change location mask for changed object candidates in the source scene. (3) 3D geometric validation is performed using SDF-derived features per object candidate for change mask refinement. We evaluate our approach on both synthetic and real-world datasets and demonstrate its effectiveness via the task of changed object detection. Supplementary video is available at: \href{https://youtu.be/oh-MQPWTwZI}{https://youtu.be/oh-MQPWTwZI}
\end{abstract}

\section{Introduction}
The ability to perform robust long-term operations is critical in many robotics and AR/VR applications, such as household cleaning and AR/VR environment scanning. Through multiple traverses of the same place, agents accumulate a more holistic understanding of their working environments. However, in the long-term setting, the working environment is prone to changes over time, e.g., the removal of a coffee mug. Conflicts may then arise when agents try to synthesize scans from different sessions. Therefore, agents are expected to first capture these changes and then obtain the up-to-date 3D reconstruction of the scene after all change conflicts have been resolved. 

An intuitive way to conduct change detection is through scene differencing between the two reconstructions of interest. Previous works on change detection leverage scene representations such as point clouds~\cite{finman2013toward,ambrucs2014meta,herbst2014toward,salas2013slam++} or Signed Distance Fields (SDF)~\cite{6162880,mccormac2018fusion++,reijgwart2019voxgraph} and perform point- or voxel-wise comparison~\cite{finman2013toward,ambrucs2014meta,herbst2014toward,fehr2017tsdf} \emph{globally} between the two scenes. To ensure that comparison is carried out at corresponding locations of the two observations, these methods demand consistent and precisely aligned reconstructions, which are hence susceptible to sensor noises and localization errors. 

\begin{figure}
    \vspace{+5pt}
    \centering
    \includegraphics[width=0.95\linewidth]{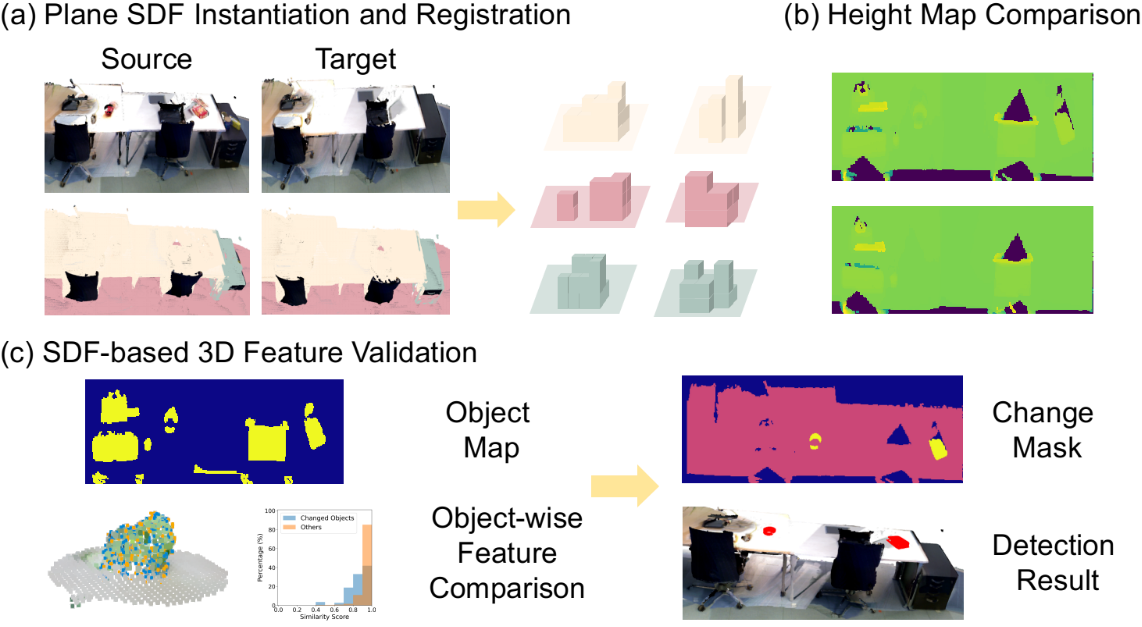}
    \caption{System Overview. Input: point clouds of the source and target scene. Output: voxels of objects detected as changes between the two scenes.  (a): For the two input point clouds, PlaneSDF volumes are fused and registered using poses of major planes (e.g., desk, cabinet, and the floor, as indicated in different colors). A 2D height map and an associated object map are obtained for each plane through projection and connected component analysis. (b): Height values for corresponding planes are compared, which yields a preliminary 2D change mask for the source plane w.r.t. the target plane. (c): The intersection of the current change mask and the source object map is found to determine changed object candidates. Each of these objects has its SDF-based features extracted and compared against the corresponding one in the target for change mask refinement. }
    \label{overview}
        \vspace{-15pt}
\end{figure}
 
We observe that most scene changes occur at the object level, and that man-made environments can often be modeled as a set of planes with objects attached to them, as opposed to a cluster of unordered points or voxels with no geometric structure. Therefore, we choose to represent the whole scene as a set of planes, each having an associated SDF volume that describes the geometric details of the objects attached to it, which we term as the PlaneSDF representation. Similiar to the idea of dividing the whole environment into submaps, e.g., based on time intervals~\cite{reijgwart2019voxgraph} or objects~\cite{mccormac2018fusion++, schmid2021panoptic}, agents could maintain multiple PlaneSDF volumes of scalable sizes in lieu of a single chunk of global SDF while saving update and memory reload time by updating volumes only in the current viewing frustum. Furthermore, this representation is also more robust to localization drift as local regional correction can be performed patch by patch each time two planes from different traverses are registered via plane pose.

Taking advantage of the PlaneSDF representation, in this paper, we propose a change detection algorithm given a source and a target scene that decomposes the original global comparison in a local plane-wise fashion. Treating each plane as a separator, the local change detection is performed plane-wise as well as at the object level. The global localization drift issue between two scenes is alleviated during plane-pose registration. Through the projection of SDF voxel height values onto the plane, the obtained height map and its value connectivity offers a solid indication about the potential object candidates along with their projected 2D contours, making it possible to conduct 3D geometric validation only on SDF voxels belonging to the potentially changed objects. Our main contributions are as follows:
\begin{enumerate}
    \item PlaneSDF is proposed as a novel representation for indoor scene reconstruction.
    \item A change detection algorithm, consisting of 2D height map comparison and 3D geometric validation, is developed leveraging the PlaneSDF data structure.
    \item The effectiveness of the proposed algorithm is demonstrated on both synthetic and real-world datasets of indoor scenes.
\end{enumerate}
\begin{figure*}
    \vspace{+5pt}
    \centering
    \includegraphics[scale=0.4]{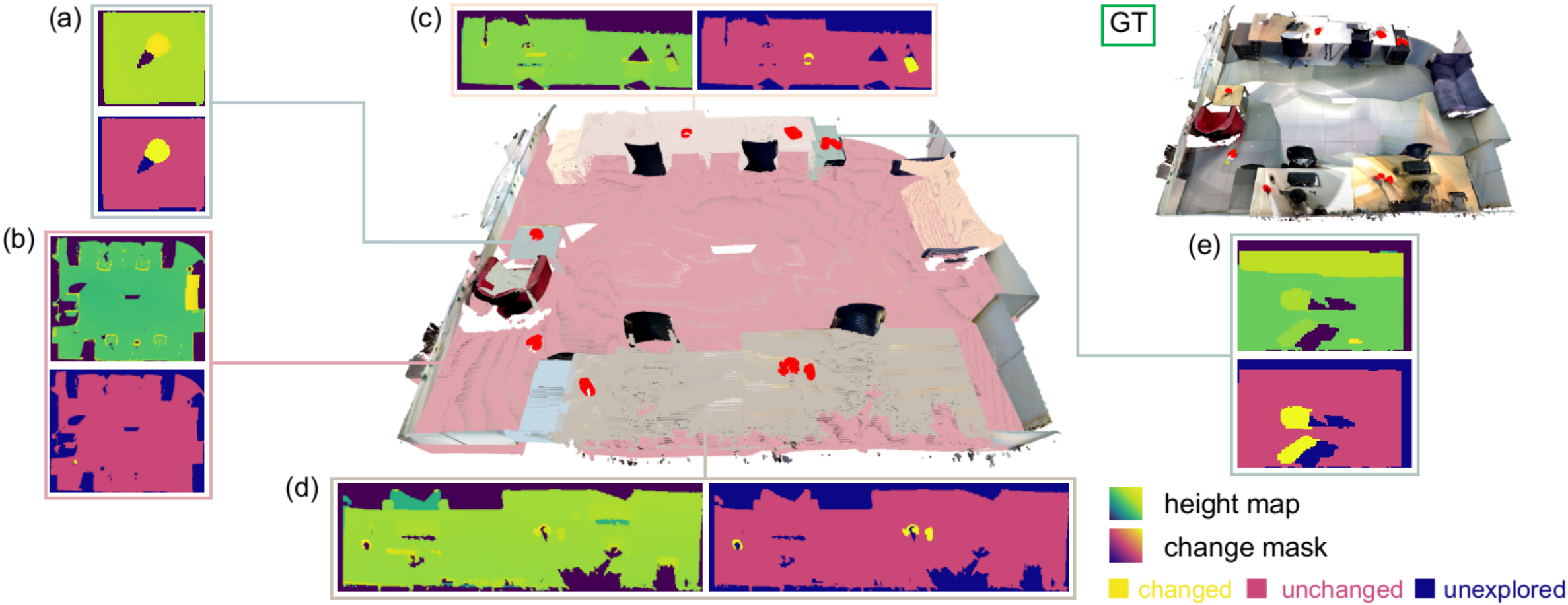}
    \caption{Change detection results for a complete indoor scene from the object change detection dataset~\cite{langer2020robust}. The whole scene is spatially subdivided into multiple PlaneSDF instances (marked by distinct colors). Note that there could be some overlap among certain SDF volumes (e.g., the seating area of the sofa in the upper right of the scene is also fused into the floor volume). For each plane of interest, i.e., planes with objects newly introduced onto them, the associated height map and the final change mask are shown. The detected object changes are colored in red while the ground truth (GT) changes are rendered in the upper right corner of the figure.  }
    \label{panorama}
    \vspace{-15pt}
\end{figure*}
\section{Related Work}
Change detection, as widely discussed in research concerning long-term robotic operations, can be roughly divided into two categories: geometric and probabilistic approaches.

\textbf{Geometric Approaches.} Geometric approaches are usually based on comparing geometric features extracted from various environment representations. Walcott-Bryant et al.~\cite{walcott2012dynamic} developed Dynamic Pose Graph SLAM, where change detection is performed on the 2D occupancy grid to edit and update the pose graph. Classical 2D feature descriptors, e.g. SURF, ORB, and BRISK~\cite{derner2019towards,derner2021change}, were extracted from the grey scale input images and the visual database, respectively. Next, the Euclidean distance between the two features is computed to determine if changes have taken place. There are also many works in the literature which use 3D representations. Finman et al.~\cite{finman2013toward} performed scene differencing on depth data among multiple maps and learned segmentation models with surface normals and color edges to discover new objects in the scene. Ambrus et al.~\cite{ambrucs2014meta} computed a meta-room reference map of the environment from the collected point cloud, and employed spatial clustering based on global descriptors to discover new objects in the scene. Fehr et al.~\cite{fehr2017tsdf} adapted volumetric differencing onto a multi-layer SDF grid and showed its effectiveness in object discovery and class recognition. Kunze et al.~\cite{kunze-a} built and updated a hierarchical map of the environment by comparing object positions between observations and corresponding map contents. Schmid et al.~\cite{schmid2021panoptic} proposed a panoptic map representation using multiple Truncated Signed Distance Fields for each panoptic entity to detect long-term object-level scene changes on-the-fly. Langer et al.~\cite{langer2020robust} combined semantic as well as supporting plane information, and conducted local verification (LV) to discover objects  newly introduced into the scene. The proposed method outperforms several global point- and voxel-based approaches and is  selected as the baseline here for comparison. 

\textbf{Probability-based Approaches.} Previous works in this category tend to develop statistical models to describe sensor measurement or environment dynamics. Krajnik et al.~\cite{krajnik2014spectral} modeled the environment's spatio-temporal dynamics by its frequency spectrum, while~\cite{herbst2014toward,luft2018detecting} exploited probabilistic measurement models to indicate how likely it is for each surface element in the scene to have moved between two scenes. Bore et al.~\cite{bore2018detection} proposed a model for object movement describing both local moves and long-distance global motion. Katsura et al.~\cite{katsura2019spatial} converted point clouds and measured data into ND (Normal Distribution) voxels using the Normal Distribution Transform (NDT) and compared voxel-wise distribution similarity. 

There are also learning-based change detection approaches~\cite{alcantarilla2018street,wald2019rio} that learn geometric features through neural networks trained on pre-registered images or SDF pairs. Considering the potential challenges of training data availability and generalization to unseen changes, this paper focuses only on non-learning based methods.

 Despite all the results reported, the global point- or voxel-wise geometric comparisons are susceptible to sensor noises and localization errors and the results of probabilistic approaches may not be readily applicable to scene mapping tasks. Hence, in this work, we consider 2D as well as 3D information on the voxel and object level with the proposed PlaneSDF structure, and achieves robust change detection on both synthetic and real-world datasets.

\section{Method Overview}
Our method (see Fig.~\ref{overview}) leverages the plane-to-object supporting structure through the PlaneSDF representation, thereby enabling us to first perform local pairwise plane pose alignment against global reconstruction errors. We then obtain change detection results via  efficient and effective local scene comparison on 2D height map and 3D object surface geometry informed by the SDF volume.

\subsection{PlaneSDF Instantiation} \label{fuseSDF}
We first generate the PlaneSDF representation for each scene, i.e., representing the input 3D point cloud for the scene as a set of planes and their associated SDF volumes.

For plane detection, when given sequential point cloud streams, we extract planes from each frame with RANSAC and merge them when a new frame arrives,  as how SLAM systems commonly proceed when using planes as pose estimation constraints~\cite{taguchi2013point,ma2016cpa,hsiao2017keyframe}. When a point cloud for the complete scene is available, we run a spatial clustering algorithm~\cite{straub2015small} to detect a set of planes out of the cloud. 

For each plane detected, we fuse an SDF volume using all the points within a predefined distance to the plane, in the hope that the obtained SDF will record the free space and object geometry solely from objects directly supported by the plane, e.g., the drawings hanging on the vertical wall or the soda can placed on the table. Note that when two detected planes are less than the defined fusing distance away from each other or there are bigger objects supported by multiple planes, a point could be fused into multiple PlaneSDF instances, e.g., the color overlap of the sofa and the floor instances in Fig.~\ref{panorama}). We also limit our detection of planes to only horizontal and vertical ones, as they constitute most of the ``plane-supporting-objects'' cases we encounter in daily lives.

Furthermore,  the local 2D height grid map evaluated w.r.t. the plane is computed, where each grid stores the maximum voxel-to-plane distance in the height direction at the current plane location. The height map is non-zero for plane locations occupied by objects, zero for flat unoccupied locations, and $-1$ for unobserved regions. Building on top of this, as non-zero regions are disconnected from each other by the plane zero-level set, we could easily obtain an ``object (or object cluster for multiple small objects close to each other) map'' (Fig.~\ref{overview}(d)) preserving relatively accurate object contours through connected component labeling on the height map.

Given two PlaneSDF volumes, a source and a target, instantiated from the two scenes respectively, we define the 2D change mask of the pair as a ternary mask of the same size as the \emph{source} height map, indicating all changed plane locations in the \emph{source} w.r.t. the \emph{target} (Fig.~\ref{panorama}).

\subsection{PlaneSDF Registration}\label{reg} 
 Before scene differencing is conducted, PlaneSDF volumes of the two scenes are first registered so that the comparison is guaranteed to be carried out on two observations of the same plane. With the assumption that input point clouds from different sessions share the same world coordinate frame, registration of PlaneSDF volumes is accomplished through plane poses to alleviate the effect of localization drift among reconstructions of the same plane. For each pair of PlaneSDFs, we determine if they belong to the same plane according to the orientation cosine similarity and offset difference of the two plane poses:
\begin{equation} 
\begin{aligned}
         &\bf{n^Tn'} \geq \delta_{\bf{n}}\\ &||d-d'||\leq \delta_d, \\
\end{aligned}
\end{equation}
where $({\bf{n}},d)$ and $({\bf{n'}} ,d')$ are the plane surface normals and offsets from origin of the source and target PlaneSDF volumes, respectively. $\delta_{\bf{n}}$ and $\delta_d$ are the minimum cosine similarity and maximum offset distance for two planes to be regarded as the same plane. In this way, via associating plane detections of similar orientations and offsets in the pair of reconstructions,  small localization drift of the same plane can be mitigated by applying the relative transform between plane poses, from which we are then ready for change detection on each registered PlaneSDF pair.

\subsection{Height Map Comparison}\label{hm}
\begin{figure}
\vspace{+5pt}
    \centering
    \includegraphics[width=0.75\linewidth]{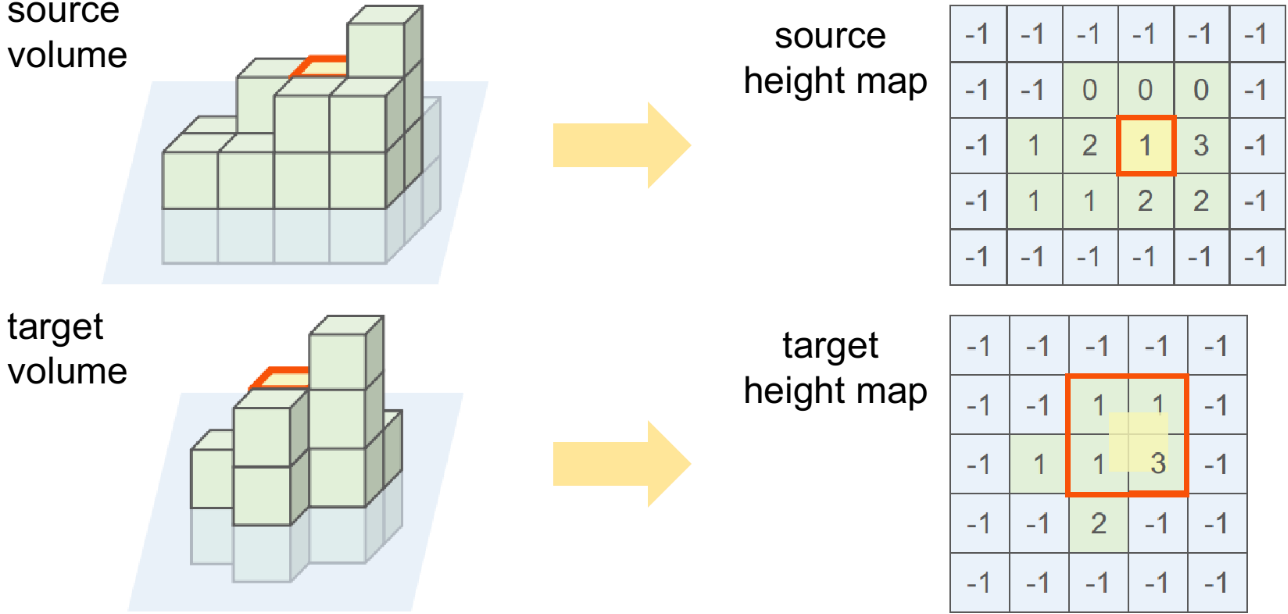}
    \caption{Height map comparison. For the registered source and target PlaneSDF pairs, each grid in the source height map is projected onto the target height map, with its height value compared against those of its closest 2$\times$2 neighborhood. If all four neighbors have a height difference above a threshold, this grid (plane location) is preliminarily marked as changed.}
    \label{hmc}
    \vspace{-15pt}
\end{figure}

As floating objects are rare in daily scenes, height value discrepancy at the same plane location in different observations can offer informative speculation about the changes on this plane, e.g., when objects are newly removed or added, drastic changes between zero and non-zero height values will occur. In this spirit, we project each location, $(x,y)$, of the source height map $H$ onto the target height map $H'$ using the relative plane pose. If the height value variation is above a threshold $\delta_h$, we mark this plane location as \emph{changed} (see Fig.~\ref{hmc}). Oftentimes, the projected location, $(x',y')$, will not not land exactly onto a grid center in the target map, so comparisons are drawn between the source height and those of the four nearest neighbors of $(x',y')$:
\begin{equation}
\begin{aligned}
   \sum\limits_{\substack{i=0,1; j=0,1}}\mathds{1}(|H'(\left\lfloor x'\right\rfloor+i,\left\lfloor y'\right\rfloor+j)-&H(x,y)| \leq \delta_h) 
   \\ = \begin{cases}0, & \text{changed}\\\geq 1, & \text{unchanged}.
\end{cases}
\end{aligned}
\end{equation}

In most cases, as a consequence of measurement noises, the change mask obtained after direct comparison is usually corrupted by small false positive clusters scattered around the map. Therefore, a round of connected component filtering followed by dilation is applied to remove the noise.

\subsection{3D Voxel Validation}\label{3d} 
Comparing height values for changes works well when (1) objects are removed or added, inducing significant variation in height values, or (2) camera trajectories have a high observation overlap of the unchanged objects between two runs. However, height implications can fail easily when old objects are replaced with new ones in the same place, or different parts of the same unchanged object are observed due to disparate viewing angles.

Therefore, 3D validation on the SDF of potential changed source plane locations is introduced with the goal of correcting false positives indicated by the change mask. For the overlapping space of two observations, if the same object persists, then the local surface geometry and free space description should be similar, or the target SDF will otherwise be remarkably different from that of the source.

Here, for the sake of selecting key voxels and obtaining corresponding descriptive geometry characterization around the selected locations, the curvature-derived description of the SDF is adopted for its capability to characterize the geometry of both object surfaces and the unoccupied space in between. In addition to indicating the planarity, convexity, or concavity of the object surface, the trend of SDF variation amid object surfaces can reflect inter-surface spatial relations, e.g., the sudden drop of an increasing SDF value along a ray direction can imply the switch of the nearest reference surface for SDF value calculation as the ray marches through surfaces. In contrast, the raw SDF value description and its gradient-derived counterpart are less suitable for the unified goal of key voxel selection and local geometry description. The former, due to the unavailability of ground truth surfaces during point fusion, is prone to slight inconsistency when constructed from different camera trajectories, while the latter returns an indistinguishable magnitude of one by construction in most places.

\begin{figure}
\vspace{+5pt}
    \centering
    \includegraphics[width=\linewidth]{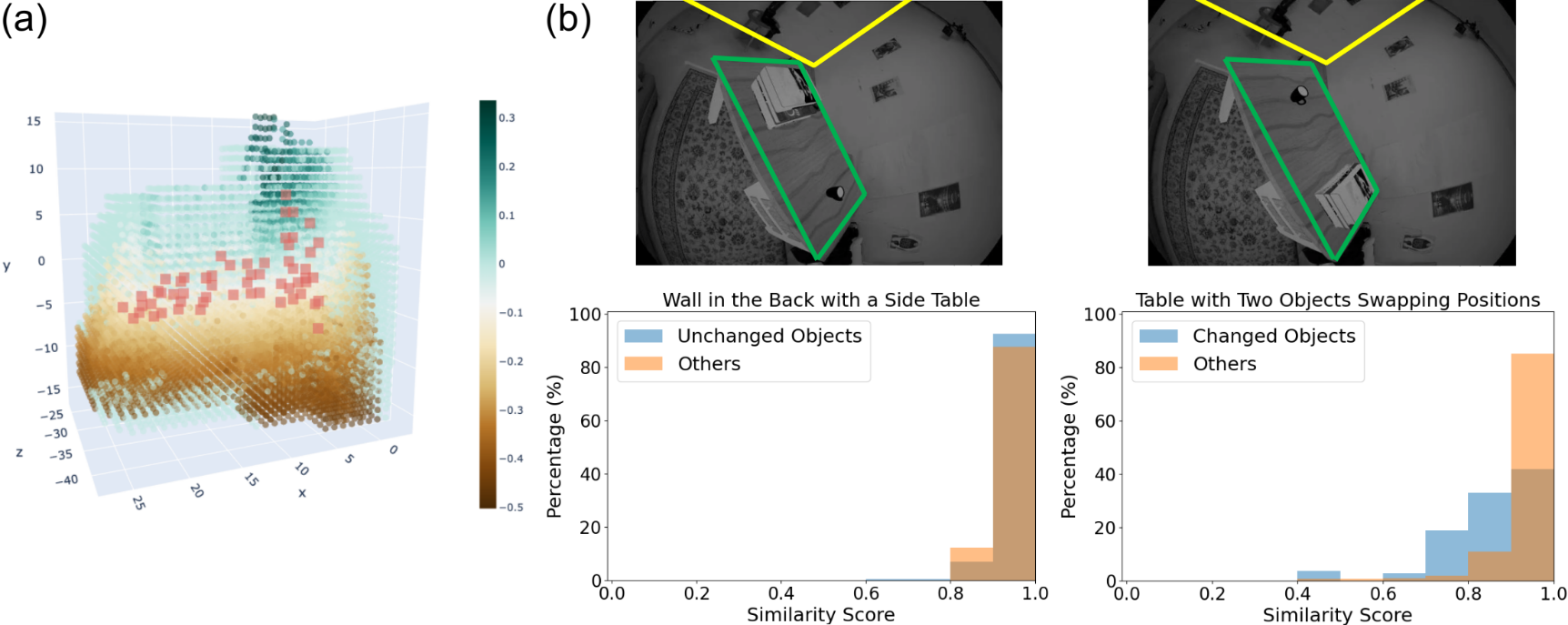}
    \caption{Key voxel distribution and corresponding similarity score distribution of planes with and without changes. (a) Key voxel (red square dots) distribution within a voxel blob (round dots with colors indicating the SDF value). (b) Key voxels within the same PlaneSDF volume are classified as either ``part of an object'' or ``others'' as everything left in the background. Left (PlaneSDF of the yellow plane): Both the side table (object) and the wall (others) are unchanged, hence both similarity scores bias towards higher-valued bins. Right (PlaneSDF of the green plane): The book stack and the coffee mug swap their positions on the table. Their shape distinction leads to scattered distribution of voxel similarity scores at the same 3D position, while the ``other" unchanged voxels around the tabletop plane still share high similarity.}
    \label{hist} 
    \vspace{-15pt}
\end{figure}

Additionally, to make the comparison more robust to measurement noises and reconstruction errors, the SDF voxels of interest are extracted and compared in the minimal unit of an object (cluster). This is achieved by selecting voxel blobs in each source PlaneSDF as those whose 2D projected clusters from the change mask have high overlap with the connected clusters in the object map, i.e., the intersection of the change mask and the object map. Through per-blob 3D geometry validation, the final change mask not only preserves a more detailed object contour in cases of adding/removing an object to/from a free space, but also self-corrects false per-voxel height variation induced by sensor noises in a clean way.

\textbf{Key Voxel Selection. } Key voxels are selected per object blob so as to offer a more compact and robust characterization of the overall blob shape. Inspired by~\cite{millane2021freetures}, voxels around regions of high curvature are selected as key voxels, implying neighborhoods of significant shape variations (see Fig.~\ref{hist}(a). We adopt the measure of local extrema of the determinant of Hessian (DoH), $det(Hess(\boldsymbol{v}))$, and calculate the Hessian matrix within a complete $3\times 3\times 3$ neighborhood $\mathcal{N}$:
\begin{equation}
\begin{aligned}
&Hess(\boldsymbol{v}) =    \begin{bmatrix}
    s_{xx} & s_{xy} & s_{xz}\\
    s_{yx} & s_{yy} & s_{yz}\\
    s_{zx} & s_{zy} & s_{zz}
    \end{bmatrix} \\
   & s_{ij} =(\textbf{G}_j*\textbf{G}_i)(\Phi(\boldsymbol{v})) \quad i,j = x,y,z,
\end{aligned}
\end{equation}
where each element $s_{ij}$ in the Hessian matrix of $\boldsymbol{v}$ is obtained via convolution of $\Phi(\boldsymbol{v})$, the $3\times 3\times 3$ SDF neighborhood at $\boldsymbol{v}$, with the 3D Sobel filter $\textbf{G}$ in turn in the $i$ and $j$ direction.  

\textbf{Per-voxel Shape Description.}
For each key voxel $\boldsymbol{v}_0$ in the object blob $\mathcal{O}$, the three eigenpairs of the Hessian matrix, $\boldsymbol{p}_i = (\lambda_i,\boldsymbol{e}_i), i=x,y,z$, are computed and represent the three principal curvatures ($\lambda_i$s) and their directions ($\boldsymbol{e}_i$s) at $\boldsymbol{v}$, respectively. This operation is then repeated for each voxel in $\mathcal{N}$ and its corresponding neighborhood $\mathcal{N}'$ in the target map (determined by its projected location $\boldsymbol{v}'$ in the target map). The three eigenvalues are normalized for numerical stability and each principal direction vector $\boldsymbol{e}_i$ is converted into spherical coordinate $(\theta_i,\phi_i)$.

We then construct eigenpair histograms, $\mathcal{H}$ and $\mathcal{H}'$, for the corresponding neighborhood $\mathcal{N}$ and $\mathcal{N}'$. For neighborhood $\mathcal{N}$, we compute three sub-histograms, $h_i$s, for all the eigenpairs  $\boldsymbol{p}_{j_i}$ in the $i$ direction, where $i=x,y,z$:
\begin{equation}
    \begin{aligned}
     \boldsymbol{p}_{j_i} &= [\theta_i,\phi_i,\lambda_i], \boldsymbol{p}_j \in \mathcal{N} \\ 
    &\Rightarrow h_i\in \mathds{R}^{N_\theta\times N_{\phi} \times N_\lambda}\\
        \theta_{h_i} & = [0,180^\circ], \phi_{h_i}=[-90^\circ,90^\circ]\\
      \lambda_{h_i}&=[\min\limits_{j\in\mathcal{N}}(\lambda_{j_i}),\max\limits_{j\in\mathcal{N}}(\lambda_{j_i})],
    \end{aligned}
\end{equation}
where $N_{\theta}$,$N_{\phi}$, and $N_{\lambda}$ are the number of bins, and $\theta_{h_i}$,$\phi_{h_i}$, and $\lambda_{h_i}$ are the bin threshold in the $\theta$, $\phi$, and $\lambda$ directions for $h_i$, respectively. With each $h_i$ of dimension $N_\theta\times N_{\phi} \times N_\lambda$, we then concatenate the three to form the final histogram, $\mathcal{H} = [h_1 \vert \vert h_2 \vert \vert h_3]$, describing the local shape distribution around this key voxel in the source. The corresponding $\mathcal{H}'$ for the target neighborhood is computed in the same fashion, while sharing all the histogram thresholds with those of $\mathcal{H}$.

To further enhance its ability of characterizing local shapes, we append the final histogram with a weighted signed distance value $s$ of the neighborhood. The weights are assigned with a Gaussian filter centered at $\boldsymbol{v_0}$ with deviation of $\sigma=2$, and the weighted SDF $s$ is computed as follows:
\begin{equation}
\begin{aligned}
      & w_i = \frac{1}{\sqrt{2\pi}\sigma}e^{\frac{(\boldsymbol{v}_i-\boldsymbol{v}_0)^2}{2\sigma^2}},\boldsymbol{v}_i\in\mathcal{N}(\boldsymbol{v}_0)\\
      & s = \frac{\sum w_i \Phi(\boldsymbol{v}_i)}{\sum w_i}.
\end{aligned}
\end{equation}
Thus the ultimate feature vector for the key voxel in the source is $f(\boldsymbol{v}_0)=[\mathcal{H},s]$, which is of dimension $3\times N_\theta\times N_{\phi} \times N_\lambda+1$. We define a similarity score, $sim\in (0,1)$, at this key voxel between the two features, $f$ and $f'$, of the source and target map, respectively, as:
\begin{equation}
    sim(f,f') = 1/(1+\alpha \lVert f-f'\rVert_2),
\end{equation}
where $f'(\boldsymbol{v}_0)=[\mathcal{H}',s']$ and $\alpha$ is a coefficient for adjusting the contribution of the Euclidean distance between $f$ and $f'$, $\lVert f-f'\rVert_2$, to the similarity score.

\textbf{Per-object Shape Comparison. }The distribution of the similarity scores for all key voxels in the current object blob then makes it possible to determine if the space is occupied by the same object across two sessions. We argue that for an \emph{unchanged} space occupied with the same object blob, the similarity scores, as an indication of the local shape, should be concentrating around higher values, whereas for a space with objects later removed, added, or replaced by another object, they should either be low (removed or added) or distributed more evenly around a wider range of bins (replaced) (see Fig.~\ref{hist}(b)). Therefore, we construct the similarity score histogram for the object blob and compute the histogram mean to determine if the object has changed:
\begin{equation}
\begin{aligned}
   & H_{avg}  = \sum\limits m_in_i/N \\
&    isChanged(\mathcal{O}) = \mathds{1}( H_{avg} < \delta_{blob}),
\end{aligned}
\end{equation}
where $m_i$ and $n_i$ are the midpoint value and frequency of each bin $i$, and $N$ is the total number of key voxels in this object blob. The object is then validated as changed if $H_{avg}$ is below a similarity threshold, $\delta_{blob}$, or false positives from 2D comparison can be corrected based on the relatively high $H_{avg}$ value.

Following the plane locations marked as changed in the change mask, all the corresponding voxels along the height direction are extracted, which are the changed part of the source scene w.r.t. the target.

\section{Experiments and Results}
In this section, we evaluate our approach on both synthetic and real-world indoor datasets, and demonstrate its strength via tasks revolving around object-level change detection.

\subsection{Datasets}
\begin{figure}
\vspace{+5pt}
    \centering
    \includegraphics[scale=0.25]{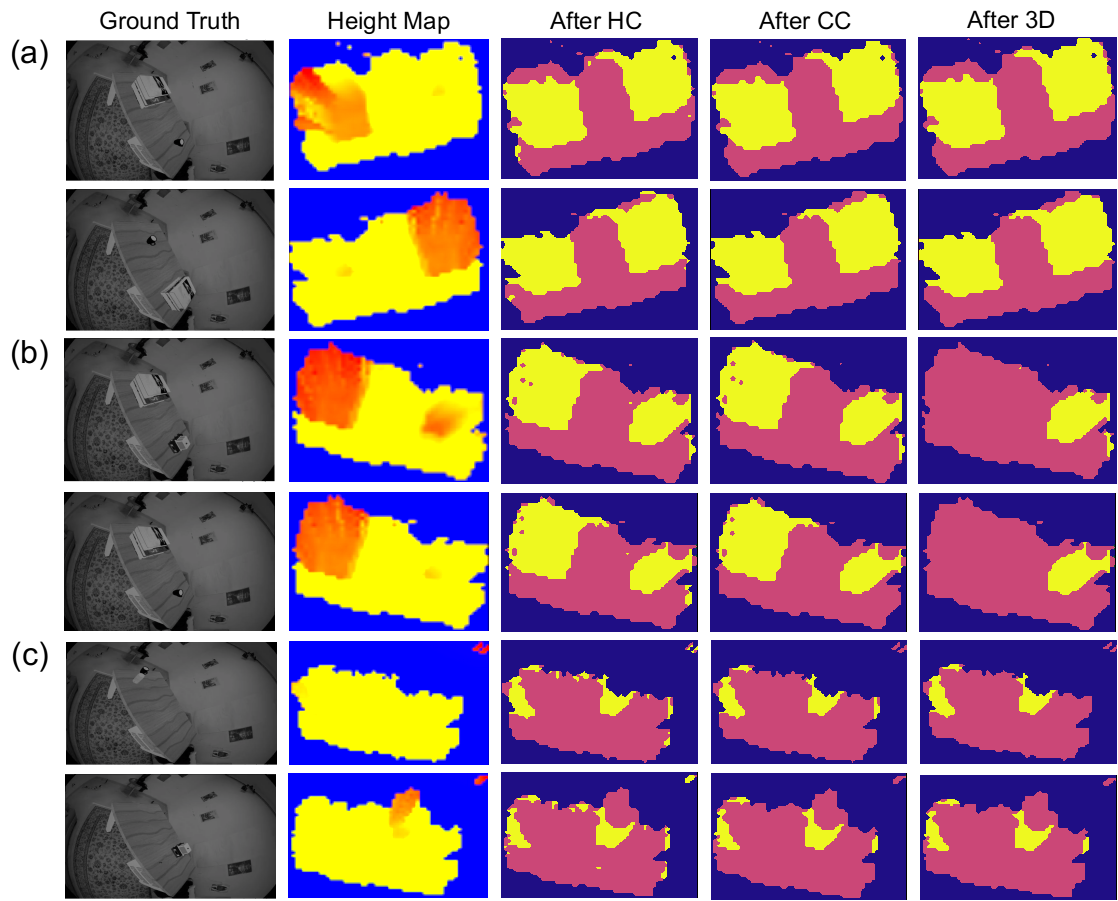}
    \caption{Sample change detection results on the synthetic tabletop dataset. Each mask showcases the change detection result of treating the sequence in the same row as the source. Here we include snapshots of the actual scene in the first column, the associated height map in the second column, and the evolution of the change mask out of each stage of our approach in the last three columns: (1) height map comparison (HC) (2) connected component filtering and dilation (CC) (3) 3D geometric validation (3D).}
    \label{sim} 
    \vspace{-15pt}
\end{figure}

\textbf{Synthetic Tabletop Dataset.}
For evaluations under controlled environments, we generated synthetic indoor sequences with known object models on a tabletop. We first scanned a static, furnished room with a Lidar scanner to obtain a ground-truth 3D point cloud of the room. A few synthetic daily objects, e.g. mug and book stack, are then arbitrarily placed on a synthetic table in the scene, which are added, removed, or moved across multiple sequences, thus creating the desired changes to be detected. The scenes are rendered by simulating cameras on the Oculus Quest 2 headset moving in a preset trajectory around the table, from which per-frame 3D point cloud observations were generated and used as the input to our algorithm.

\textbf{Object Change Detection Dataset.} 
The object change detection dataset~\cite{langer2020robust} is recorded with an Asus Xtion PRO Live RGB-D camera mounted on an HSR robot, consisting of multiple complete or partial point clouds of five scenes: big room, small room, kitchen, office, and living room. Each scene consists of a reference reconstruction and 5 to 6 other reconstructions obtained using Voxblox~\cite{oleynikova2017voxblox}, 
accompanied by various levels of permanent structure misalignment and noisy boundaries due to localization and reconstruction errors. Ground truth annotation of 3 to 18 newly introduced YCB~\cite{calli2017yale} objects to the scene is provided. 

\subsection{Evaluation Metrics}
We adopt the commonly used precision and recall rates as the metrics for change detection evaluation.

For the object change detection dataset, following the measures in~\cite{langer2020robust}, we compute precision, recall rate, and F1 score at the \emph{point} level, based on the ground truth changed point annotation and our detection results.  Precision is computed as the proportion of total number of detected points that correspond to the ground truth, and recall rate is defined as the proportion of ground truth points that are incorporated in the detection points. The F1 score provides the harmonic mean of the two metrics. Two other metrics, the number of missing objects (changed objects with no points detected as changed) and wrongly detected clusters (clusters generated by the method that do not overlap with any changed objects) are also reported so as to better manifest the approach's performance on the object/cluster level. 

\begin{table*}[]
\vspace{+5pt}
\centering
\caption{Result comparison of the proposed approach with three baselines provided by the object change detection dataset. Best values are marked in bold.
(Pr = precision, Re = recall, F1 = F1 score, M = missed objects, W = wrongly detected clusters)}
\label{result}
\resizebox{\textwidth}{!}{
\begin{tabular}{|
>{\columncolor[HTML]{FFFFFF}}c |
>{\columncolor[HTML]{FFFFFF}}c 
>{\columncolor[HTML]{FFFFFF}}c 
>{\columncolor[HTML]{FFFFFF}}c 
>{\columncolor[HTML]{FFFFFF}}c 
>{\columncolor[HTML]{FFFFFF}}c |
>{\columncolor[HTML]{FFFFFF}}c 
>{\columncolor[HTML]{FFFFFF}}c 
>{\columncolor[HTML]{FFFFFF}}c 
>{\columncolor[HTML]{FFFFFF}}c 
>{\columncolor[HTML]{FFFFFF}}c |}
\hline
{\color[HTML]{000000} } & \multicolumn{5}{c|}{\cellcolor[HTML]{FFFFFF}{\color[HTML]{000000} Small Room}} & \multicolumn{5}{c|}{\cellcolor[HTML]{FFFFFF}{\color[HTML]{000000} Big Room}} \\ \hline
{\color[HTML]{000000} } & \multicolumn{1}{c|}{\cellcolor[HTML]{FFFFFF}{\color[HTML]{000000} Pr}} & \multicolumn{1}{c|}{\cellcolor[HTML]{FFFFFF}{\color[HTML]{000000} Re}} & \multicolumn{1}{c|}{\cellcolor[HTML]{FFFFFF}{\color[HTML]{000000} F1}} & \multicolumn{1}{c|}{\cellcolor[HTML]{FFFFFF}{\color[HTML]{000000} M}} & {\color[HTML]{000000} W} & \multicolumn{1}{c|}{\cellcolor[HTML]{FFFFFF}{\color[HTML]{000000} Pr}} & \multicolumn{1}{c|}{\cellcolor[HTML]{FFFFFF}{\color[HTML]{000000} Re}} & \multicolumn{1}{c|}{\cellcolor[HTML]{FFFFFF}{\color[HTML]{000000} F1}} & \multicolumn{1}{c|}{\cellcolor[HTML]{FFFFFF}{\color[HTML]{000000} M}} & {\color[HTML]{000000} W} \\ \hline
{\color[HTML]{000000} Octomap~\cite{langer2017fly}} & \multicolumn{1}{c|}{\cellcolor[HTML]{FFFFFF}{\color[HTML]{000000} 0.11±0.05}} & \multicolumn{1}{c|}{\cellcolor[HTML]{FFFFFF}{\color[HTML]{000000} 0.61±0.18}} & \multicolumn{1}{c|}{\cellcolor[HTML]{FFFFFF}{\color[HTML]{000000} 0.19±0.08}} & \multicolumn{1}{c|}{\cellcolor[HTML]{FFFFFF}{\color[HTML]{000000} 15}} & {\color[HTML]{000000} 176} & \multicolumn{1}{c|}{\cellcolor[HTML]{FFFFFF}{\color[HTML]{000000} 0.07±0.04}} & \multicolumn{1}{c|}{\cellcolor[HTML]{FFFFFF}{\color[HTML]{000000} 0.42±0.15}} & \multicolumn{1}{c|}{\cellcolor[HTML]{FFFFFF}{\color[HTML]{000000} 0.12±0.07}} & \multicolumn{1}{c|}{\cellcolor[HTML]{FFFFFF}{\color[HTML]{000000} 42}} & {\color[HTML]{000000} 434} \\ \hline
{\color[HTML]{000000} Meta-room~\cite{ambrucs2014meta}} & \multicolumn{1}{c|}{\cellcolor[HTML]{FFFFFF}{\color[HTML]{000000} 0.04±0.03}} & \multicolumn{1}{c|}{\cellcolor[HTML]{FFFFFF}{\color[HTML]{000000} 0.44±0.08}} & \multicolumn{1}{c|}{\cellcolor[HTML]{FFFFFF}{\color[HTML]{000000} 0.07±0.04}} & \multicolumn{1}{c|}{\cellcolor[HTML]{FFFFFF}{\color[HTML]{000000} 24}} & {\color[HTML]{000000} 276} & \multicolumn{1}{c|}{\cellcolor[HTML]{FFFFFF}{\color[HTML]{000000} 0.24±0.30}} & \multicolumn{1}{c|}{\cellcolor[HTML]{FFFFFF}{\color[HTML]{000000} 0.55±0.05}} & \multicolumn{1}{c|}{\cellcolor[HTML]{FFFFFF}{\color[HTML]{000000} 0.25±0.27}} & \multicolumn{1}{c|}{\cellcolor[HTML]{FFFFFF}{\color[HTML]{000000} 31}} & {\color[HTML]{000000} 464} \\ \hline
{\color[HTML]{000000} Best of~\cite{langer2020robust}} & \multicolumn{1}{c|}{\cellcolor[HTML]{FFFFFF}{\color[HTML]{000000} \textbf{0.55±0.36}}} & \multicolumn{1}{c|}{\cellcolor[HTML]{FFFFFF}{\color[HTML]{000000} 0.66±0.17}} & \multicolumn{1}{c|}{\cellcolor[HTML]{FFFFFF}{\color[HTML]{000000} 0.57±0.22}} & \multicolumn{1}{c|}{\cellcolor[HTML]{FFFFFF}{\color[HTML]{000000} \textbf{6}}} & {\color[HTML]{000000} 28} & \multicolumn{1}{c|}{\cellcolor[HTML]{FFFFFF}{\color[HTML]{000000} 0.78±0.13}} & \multicolumn{1}{c|}{\cellcolor[HTML]{FFFFFF}{\color[HTML]{000000} 0.78±0.04}} & \multicolumn{1}{c|}{\cellcolor[HTML]{FFFFFF}{\color[HTML]{000000} 0.69±0.10}} & \multicolumn{1}{c|}{\cellcolor[HTML]{FFFFFF}{\color[HTML]{000000} \textbf{2}}} & {\color[HTML]{000000} 50} \\ \hline
{\color[HTML]{000000} FPFH} & \multicolumn{1}{c|}{\cellcolor[HTML]{FFFFFF}{\color[HTML]{000000} 0.13±0.14}} & \multicolumn{1}{c|}{\cellcolor[HTML]{FFFFFF}{\color[HTML]{000000} 0.12±0.05}} & \multicolumn{1}{c|}{\cellcolor[HTML]{FFFFFF}{\color[HTML]{000000} 0.11±0.08}} & \multicolumn{1}{c|}{\cellcolor[HTML]{FFFFFF}{\color[HTML]{000000} 32}} & {\color[HTML]{000000} -} & \multicolumn{1}{c|}{\cellcolor[HTML]{FFFFFF}{\color[HTML]{000000} 0.13±0.12}} & \multicolumn{1}{c|}{\cellcolor[HTML]{FFFFFF}{\color[HTML]{000000} 0.39±0.12}} & \multicolumn{1}{c|}{\cellcolor[HTML]{FFFFFF}{\color[HTML]{000000} 0.18±0.14}} & \multicolumn{1}{c|}{\cellcolor[HTML]{FFFFFF}{\color[HTML]{000000} 19}} & {\color[HTML]{000000} -} \\ \hline
{\color[HTML]{000000} Ours} & \multicolumn{1}{c|}{\cellcolor[HTML]{FFFFFF}{\color[HTML]{000000} 0.50±0.24}} & \multicolumn{1}{c|}{\cellcolor[HTML]{FFFFFF}{\color[HTML]{000000} \textbf{0.83±0.14}}} & \multicolumn{1}{c|}{\cellcolor[HTML]{FFFFFF}{\color[HTML]{000000} \textbf{0.59±0.21}}} & \multicolumn{1}{c|}{\cellcolor[HTML]{FFFFFF}{\color[HTML]{000000} 10}} & {\color[HTML]{000000} \textbf{18}} & \multicolumn{1}{c|}{\cellcolor[HTML]{FFFFFF}{\color[HTML]{000000} \textbf{0.78±0.03}}} & \multicolumn{1}{c|}{\cellcolor[HTML]{FFFFFF}{\color[HTML]{000000} \textbf{0.85±0.15}}} & \multicolumn{1}{c|}{\cellcolor[HTML]{FFFFFF}{\color[HTML]{000000} \textbf{0.81±0.09}}} & \multicolumn{1}{c|}{\cellcolor[HTML]{FFFFFF}{\color[HTML]{000000} 8}} & {\color[HTML]{000000} \textbf{15}} \\ \hline
{\color[HTML]{000000} } & \multicolumn{5}{c|}{\cellcolor[HTML]{FFFFFF}{\color[HTML]{000000} Living Room (partial)}} & \multicolumn{5}{c|}{\cellcolor[HTML]{FFFFFF}{\color[HTML]{000000} Office (partial)}} \\ \hline
{\color[HTML]{000000} } & \multicolumn{1}{c|}{\cellcolor[HTML]{FFFFFF}{\color[HTML]{000000} Pr}} & \multicolumn{1}{c|}{\cellcolor[HTML]{FFFFFF}{\color[HTML]{000000} Re}} & \multicolumn{1}{c|}{\cellcolor[HTML]{FFFFFF}{\color[HTML]{000000} F1}} & \multicolumn{1}{c|}{\cellcolor[HTML]{FFFFFF}{\color[HTML]{000000} M}} & {\color[HTML]{000000} W} & \multicolumn{1}{c|}{\cellcolor[HTML]{FFFFFF}{\color[HTML]{000000} Pr}} & \multicolumn{1}{c|}{\cellcolor[HTML]{FFFFFF}{\color[HTML]{000000} Re}} & \multicolumn{1}{c|}{\cellcolor[HTML]{FFFFFF}{\color[HTML]{000000} F1}} & \multicolumn{1}{c|}{\cellcolor[HTML]{FFFFFF}{\color[HTML]{000000} M}} & {\color[HTML]{000000} W} \\ \hline
{\color[HTML]{000000} Octomap~\cite{langer2017fly}} & \multicolumn{1}{c|}{\cellcolor[HTML]{FFFFFF}{\color[HTML]{000000} 0.11±0.08}} & \multicolumn{1}{c|}{\cellcolor[HTML]{FFFFFF}{\color[HTML]{000000} 0.50±0.08}} & \multicolumn{1}{c|}{\cellcolor[HTML]{FFFFFF}{\color[HTML]{000000} 0.17±0.10}} & \multicolumn{1}{c|}{\cellcolor[HTML]{FFFFFF}{\color[HTML]{000000} 19}} & {\color[HTML]{000000} 74} & \multicolumn{1}{c|}{\cellcolor[HTML]{FFFFFF}{\color[HTML]{000000} 0.18±0.07}} & \multicolumn{1}{c|}{\cellcolor[HTML]{FFFFFF}{\color[HTML]{000000} 0.77±0.13}} & \multicolumn{1}{c|}{\cellcolor[HTML]{FFFFFF}{\color[HTML]{000000} 0.28±0.10}} & \multicolumn{1}{c|}{\cellcolor[HTML]{FFFFFF}{\color[HTML]{000000} 8}} & {\color[HTML]{000000} 73} \\ \hline
{\color[HTML]{000000} Meta-room~\cite{ambrucs2014meta}} & \multicolumn{1}{c|}{\cellcolor[HTML]{FFFFFF}{\color[HTML]{000000} 0.13±0.18}} & \multicolumn{1}{c|}{\cellcolor[HTML]{FFFFFF}{\color[HTML]{000000} 0.42±0.10}} & \multicolumn{1}{c|}{\cellcolor[HTML]{FFFFFF}{\color[HTML]{000000} 0.14±0.14}} & \multicolumn{1}{c|}{\cellcolor[HTML]{FFFFFF}{\color[HTML]{000000} 15}} & {\color[HTML]{000000} 122} & \multicolumn{1}{c|}{\cellcolor[HTML]{FFFFFF}{\color[HTML]{000000} 0.17±0.25}} & \multicolumn{1}{c|}{\cellcolor[HTML]{FFFFFF}{\color[HTML]{000000} 0.39±0.20}} & \multicolumn{1}{c|}{\cellcolor[HTML]{FFFFFF}{\color[HTML]{000000} 0.17±0.18}} & \multicolumn{1}{c|}{\cellcolor[HTML]{FFFFFF}{\color[HTML]{000000} 12}} & {\color[HTML]{000000} 146} \\ \hline
{\color[HTML]{000000} Best of~\cite{langer2020robust}} & \multicolumn{1}{c|}{\cellcolor[HTML]{FFFFFF}{\color[HTML]{000000} \textbf{0.83±0.29}}} & \multicolumn{1}{c|}{\cellcolor[HTML]{FFFFFF}{\color[HTML]{000000} 0.69±0.11}} & \multicolumn{1}{c|}{\cellcolor[HTML]{FFFFFF}{\color[HTML]{000000} 0.72±0.17}} & \multicolumn{1}{c|}{\cellcolor[HTML]{FFFFFF}{\color[HTML]{000000} \textbf{4}}} & {\color[HTML]{000000} \textbf{13}} & \multicolumn{1}{c|}{\cellcolor[HTML]{FFFFFF}{\color[HTML]{000000} 0.49±0.27}} & \multicolumn{1}{c|}{\cellcolor[HTML]{FFFFFF}{\color[HTML]{000000} 0.83±0.06}} & \multicolumn{1}{c|}{\cellcolor[HTML]{FFFFFF}{\color[HTML]{000000} 0.54±0.20}} & \multicolumn{1}{c|}{\cellcolor[HTML]{FFFFFF}{\color[HTML]{000000} \textbf{0}}} & {\color[HTML]{000000} 16} \\ \hline
{\color[HTML]{000000} FPFH} & \multicolumn{1}{c|}{\cellcolor[HTML]{FFFFFF}{\color[HTML]{000000} 0.11±0.11}} & \multicolumn{1}{c|}{\cellcolor[HTML]{FFFFFF}{\color[HTML]{000000} 0.31±0.14}} & \multicolumn{1}{c|}{\cellcolor[HTML]{FFFFFF}{\color[HTML]{000000} 0.15±0.14}} & \multicolumn{1}{c|}{\cellcolor[HTML]{FFFFFF}{\color[HTML]{000000} 12}} & {\color[HTML]{000000} -} & \multicolumn{1}{c|}{\cellcolor[HTML]{FFFFFF}{\color[HTML]{000000} 0.21±0.13}} & \multicolumn{1}{c|}{\cellcolor[HTML]{FFFFFF}{\color[HTML]{000000} 0.50±0.19}} & \multicolumn{1}{c|}{\cellcolor[HTML]{FFFFFF}{\color[HTML]{000000} 0.27±0.13}} & \multicolumn{1}{c|}{\cellcolor[HTML]{FFFFFF}{\color[HTML]{000000} 5}} & {\color[HTML]{000000} -} \\ \hline
{\color[HTML]{000000} Ours} & \multicolumn{1}{c|}{\cellcolor[HTML]{FFFFFF}{\color[HTML]{000000} 0.80±0.05}} & \multicolumn{1}{c|}{\cellcolor[HTML]{FFFFFF}{\color[HTML]{000000} \textbf{0.87±0.10}}} & \multicolumn{1}{c|}{\cellcolor[HTML]{FFFFFF}{\color[HTML]{000000} \textbf{0.83±0.05}}} & \multicolumn{1}{c|}{\cellcolor[HTML]{FFFFFF}{\color[HTML]{000000} \textbf{4}}} & {\color[HTML]{000000} \textbf{13}} & \multicolumn{1}{c|}{\cellcolor[HTML]{FFFFFF}{\color[HTML]{000000} \textbf{0.72±0.10}}} & \multicolumn{1}{c|}{\cellcolor[HTML]{FFFFFF}{\color[HTML]{000000} \textbf{0.94±0.08}}} & \multicolumn{1}{c|}{\cellcolor[HTML]{FFFFFF}{\color[HTML]{000000} \textbf{0.79±0.06}}} & \multicolumn{1}{c|}{\cellcolor[HTML]{FFFFFF}{\color[HTML]{000000} \textbf{0}}} & {\color[HTML]{000000} \textbf{4}} \\ \hline
{\color[HTML]{000000} } & \multicolumn{5}{c|}{\cellcolor[HTML]{FFFFFF}{\color[HTML]{000000} Kitchen (partial)}} & \multicolumn{5}{c|}{\cellcolor[HTML]{FFFFFF}{\color[HTML]{000000} Average}} \\ \hline
{\color[HTML]{000000} } & \multicolumn{1}{c|}{\cellcolor[HTML]{FFFFFF}{\color[HTML]{000000} Pr}} & \multicolumn{1}{c|}{\cellcolor[HTML]{FFFFFF}{\color[HTML]{000000} Re}} & \multicolumn{1}{c|}{\cellcolor[HTML]{FFFFFF}{\color[HTML]{000000} F1}} & \multicolumn{1}{c|}{\cellcolor[HTML]{FFFFFF}{\color[HTML]{000000} M}} & {\color[HTML]{000000} W} & \multicolumn{1}{c|}{\cellcolor[HTML]{FFFFFF}{\color[HTML]{000000} Pr}} & \multicolumn{1}{c|}{\cellcolor[HTML]{FFFFFF}{\color[HTML]{000000} Re}} & \multicolumn{1}{c|}{\cellcolor[HTML]{FFFFFF}{\color[HTML]{000000} F1}} & \multicolumn{1}{c|}{\cellcolor[HTML]{FFFFFF}{\color[HTML]{000000} M}} & {\color[HTML]{000000} W} \\ \hline
{\color[HTML]{000000} Octomap~\cite{langer2017fly}} & \multicolumn{1}{c|}{\cellcolor[HTML]{FFFFFF}{\color[HTML]{000000} 0.43±0.08}} & \multicolumn{1}{c|}{\cellcolor[HTML]{FFFFFF}{\color[HTML]{000000} 0.41±0.08}} & \multicolumn{1}{c|}{\cellcolor[HTML]{FFFFFF}{\color[HTML]{000000} 0.41±0.07}} & \multicolumn{1}{c|}{\cellcolor[HTML]{FFFFFF}{\color[HTML]{000000} 9}} & {\color[HTML]{000000} 40} & \multicolumn{1}{c|}{\cellcolor[HTML]{FFFFFF}{\color[HTML]{000000} 0.18±0.14}} & \multicolumn{1}{c|}{\cellcolor[HTML]{FFFFFF}{\color[HTML]{000000} 0.54±0.18}} & \multicolumn{1}{c|}{\cellcolor[HTML]{FFFFFF}{\color[HTML]{000000} 0.23±0.13}} & \multicolumn{1}{c|}{\cellcolor[HTML]{FFFFFF}{\color[HTML]{000000} 18.6}} & {\color[HTML]{000000} 159.4} \\ \hline
{\color[HTML]{000000} Meta-room~\cite{ambrucs2014meta}} & \multicolumn{1}{c|}{\cellcolor[HTML]{FFFFFF}{\color[HTML]{000000} 0.56±0.17}} & \multicolumn{1}{c|}{\cellcolor[HTML]{FFFFFF}{\color[HTML]{000000} 0.35±0.12}} & \multicolumn{1}{c|}{\cellcolor[HTML]{FFFFFF}{\color[HTML]{000000} 0.44±0.14}} & \multicolumn{1}{c|}{\cellcolor[HTML]{FFFFFF}{\color[HTML]{000000} 9}} & {\color[HTML]{000000} 70} & \multicolumn{1}{c|}{\cellcolor[HTML]{FFFFFF}{\color[HTML]{000000} 0.23±0.26}} & \multicolumn{1}{c|}{\cellcolor[HTML]{FFFFFF}{\color[HTML]{000000} 0.43±0.13}} & \multicolumn{1}{c|}{\cellcolor[HTML]{FFFFFF}{\color[HTML]{000000} 0.21±0.20}} & \multicolumn{1}{c|}{\cellcolor[HTML]{FFFFFF}{\color[HTML]{000000} 18.2}} & {\color[HTML]{000000} 215.4} \\ \hline
{\color[HTML]{000000} Best of~\cite{langer2020robust}} & \multicolumn{1}{c|}{\cellcolor[HTML]{FFFFFF}{\color[HTML]{000000} 0.62±0.21}} & \multicolumn{1}{c|}{\cellcolor[HTML]{FFFFFF}{\color[HTML]{000000} \textbf{0.92±0.07}}} & \multicolumn{1}{c|}{\cellcolor[HTML]{FFFFFF}{\color[HTML]{000000} 0.55±0.11}} & \multicolumn{1}{c|}{\cellcolor[HTML]{FFFFFF}{\color[HTML]{000000} \textbf{0}}} & {\color[HTML]{000000} 55} & \multicolumn{1}{c|}{\cellcolor[HTML]{FFFFFF}{\color[HTML]{000000} 0.64±0.27}} & \multicolumn{1}{c|}{\cellcolor[HTML]{FFFFFF}{\color[HTML]{000000} 0.74±0.14}} & \multicolumn{1}{c|}{\cellcolor[HTML]{FFFFFF}{\color[HTML]{000000} 0.61±0.16}} & \multicolumn{1}{c|}{\cellcolor[HTML]{FFFFFF}{\color[HTML]{000000} \textbf{2.8}}} & {\color[HTML]{000000} 34.2} \\ \hline
{\color[HTML]{000000} FPFH} & \multicolumn{1}{c|}{\cellcolor[HTML]{FFFFFF}{\color[HTML]{000000} 0.57±0.16}} & \multicolumn{1}{c|}{\cellcolor[HTML]{FFFFFF}{\color[HTML]{000000} 0.62±0.11}} & \multicolumn{1}{c|}{\cellcolor[HTML]{FFFFFF}{\color[HTML]{000000} 0.59±0.14}} & \multicolumn{1}{c|}{\cellcolor[HTML]{FFFFFF}{\color[HTML]{000000} 4}} & {\color[HTML]{000000} -} & \multicolumn{1}{c|}{\cellcolor[HTML]{FFFFFF}{\color[HTML]{000000} 0.22±0.21}} & \multicolumn{1}{c|}{\cellcolor[HTML]{FFFFFF}{\color[HTML]{000000} 0.38±0.21}} & \multicolumn{1}{c|}{\cellcolor[HTML]{FFFFFF}{\color[HTML]{000000} 0.26±0.21}} & \multicolumn{1}{c|}{\cellcolor[HTML]{FFFFFF}{\color[HTML]{000000} 14.6}} & {\color[HTML]{000000} -} \\ \hline
{\color[HTML]{000000} Ours} & \multicolumn{1}{c|}{\cellcolor[HTML]{FFFFFF}{\color[HTML]{000000} \textbf{0.77±0.015}}} & \multicolumn{1}{c|}{\cellcolor[HTML]{FFFFFF}{\color[HTML]{000000} 0.85±0.05}} & \multicolumn{1}{c|}{\cellcolor[HTML]{FFFFFF}{\color[HTML]{000000} \textbf{0.81±0.03}}} & \multicolumn{1}{c|}{\cellcolor[HTML]{FFFFFF}{\color[HTML]{000000} 2}} & {\color[HTML]{000000} \textbf{3}} & \multicolumn{1}{c|}{\cellcolor[HTML]{FFFFFF}{\color[HTML]{000000} \textbf{0.72±0.16}}} & \multicolumn{1}{c|}{\cellcolor[HTML]{FFFFFF}{\color[HTML]{000000} \textbf{0.86±0.12}}} & \multicolumn{1}{c|}{\cellcolor[HTML]{FFFFFF}{\color[HTML]{000000} \textbf{0.76±0.13}}} & \multicolumn{1}{c|}{\cellcolor[HTML]{FFFFFF}{\color[HTML]{000000} 4.8}} & {\color[HTML]{000000} \textbf{10.6}} \\ \hline
\end{tabular}
}
\vspace{-15pt}
\end{table*}

\begin{figure}
\vspace{+5pt}
    \centering
    \includegraphics[scale=0.4]{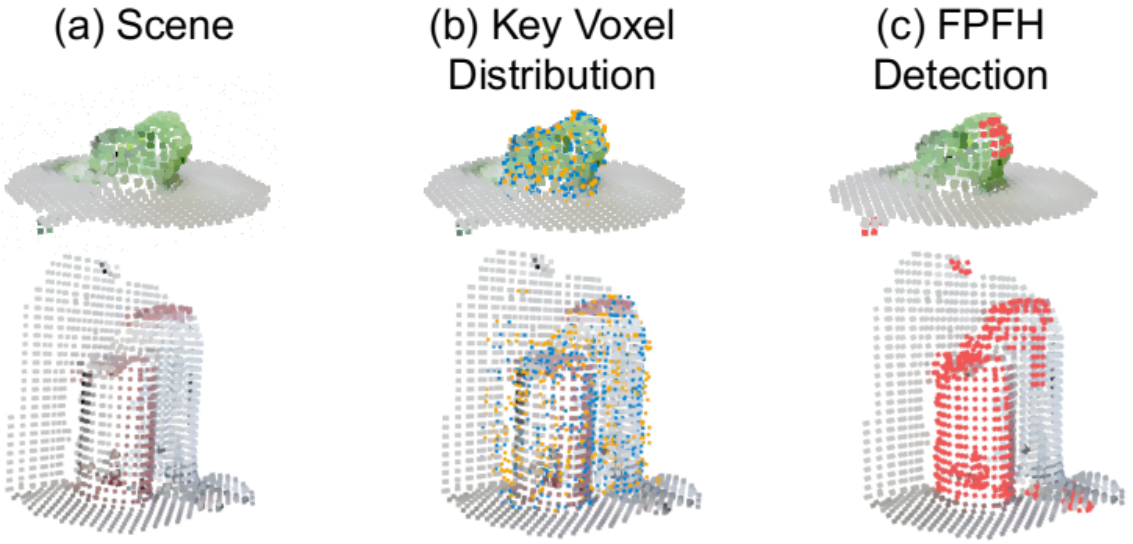}
    \caption{Illustration of key voxel distribution and detected false positive points from the per point FPFH feature matching baseline for two unchanged sub-scenes. (a) Scene rendering. Above: an isolated green object in the center of a tabletop in the ``small room'' scene. Below: two bottles standing together against a wall in the ``kitchen'' scene. (b) The scene point clouds are rendered in bigger colored squares, and the key voxels are in smaller squares with blue ones as those near object surfaces and orange ones farther away in the unoccupied space amid object surfaces. (c) Falsely detected changed points from the FPFH feature matching baseline are rendered in red.}
    \label{keypts} 
    \vspace{-15pt}
\end{figure}
\subsection{Implementation Details}
 We follow the procedures described in \ref{fuseSDF} for generating PlaneSDF instances, with the RANSAC-based approach for data streams of the synthetic tabletop dataset and the clustering-based approach for scene point clouds of the object change detection dataset. We set the fusing threshold to include points within 0.3m from the plane, hoping to cover most of the easy-to-move daily objects supported by a plane. The SDF voxel grid resolution is set as 7mm so as to best preserve the scene geometry, especially for smaller objects.

For PlaneSDF registration, the minimum cosine similarity and maximum offset  distance are set as $\delta_{\bf n} = 0.95$ and $\delta_d =0.2$m.
\begin{figure}
\vspace{+5pt}
    \centering
    \includegraphics[scale=0.34]{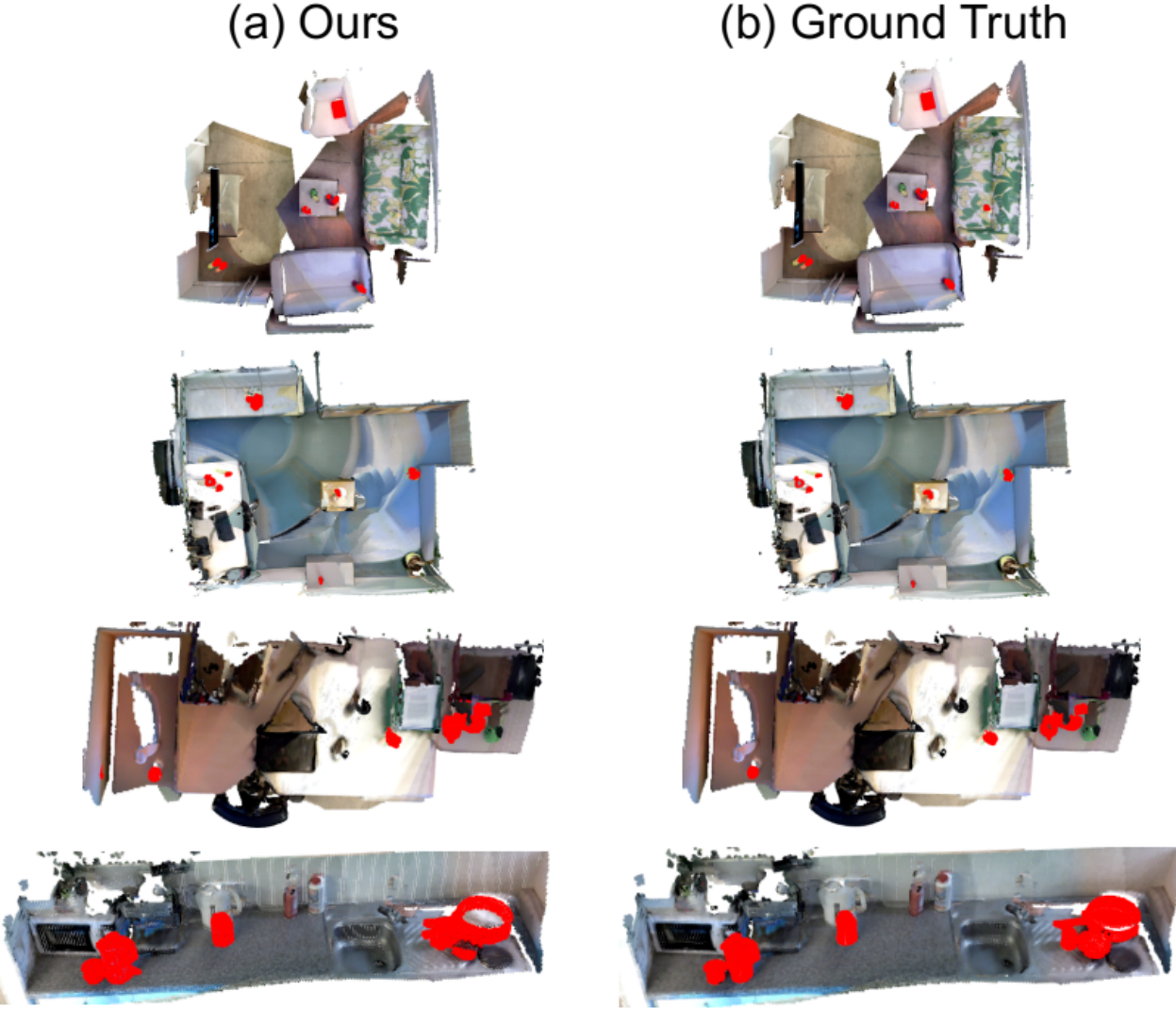}
    \caption{Qualitative examples of the change detection results (red) for the four scenes in the object change detection dataset, from top to bottom: living room (partial), small room, office (partial), kitchen (partial). (a): Detected objects from our algorithm. (b) Ground truth. }
    \label{compare} 
    \vspace{-15pt}
\end{figure}
For change detection, the height map difference threshold is set to be $\delta_h=0.02$m so as to not miss smaller objects. To construct the 3D feature histogram for each area of interest, the number of bins along each dimension is set to be $N_\phi=5$, $N_\theta=5$, $N_\lambda=6$. The $\alpha$ and the threshold $\delta_{blob}$ are set as $(\alpha,\delta_{blob}) = (2,0.9)$ for the synthetic dataset, and further moderately tuned for the object change detection dataset to accommodate certain dataset-defined cases where some slightly moved planes are not marked as changed.

\subsection{Results on the Synthetic Tabletop Dataset}
The tabletop dataset captures a relatively complete surrounding view of the various objects on a tabletop, which provides a simple yet effective scene for initial evaluation of the proposed algorithm. The experiments are run on 20 arbitrarily selected source-target sequence pairs, with objects on the tabletop ranging from coffee mug (5- cm in height), toy car (10 cm in height), to  3-layer book stack (30+ cm in height), etc. The output is the 2D change mask of the same size as the height map of the source PlaneSDF volume, indicating all the changed locations on the source plane w.r.t. the target. To prove the robustness of our algorithm, we also run all the experiments in a bi-directional fashion, i.e., detecting changes source-to-target as well as target-to-source.

With relatively complete observation of all the tabletop objects, for the 20 pairs we have tested, the algorithm is able to achieve 100\% recall and 80\% precision rate for detecting changed objects without 3D geometric validation. The precision rate further rises to 100\% after incorporating 3D validation, where false positive height differences are corrected by verifying the shape similarity in the SDF field (as for the case of the book stack shown in Fig.~\ref{sim}(b)). 

Fig.~\ref{sim} shows examples of the evolution of change masks out of each stage in the proposed method for three common object changing scenarios: (a) Two objects swap places. (b) One object changes and one remains. (c) Objects are added/removed to/from a free space. We can see that the masks out of height map comparison ($3^{rd}$  column) still contains noisy false positive (FP) clusters, as a consequence of reconstruction errors. The smaller FP clusters are then partially removed by connected-component filtering and dilation, as shown in the $4^{th}$ column, but bigger FP patches still persist, such as the book stack on the left side of the tabletop in scenario (b). The 3D validation here then plays a significant role in comparing the 3D geometric similarity of all the possible patches and effectively reverting the FP book stack back to unchanged ($5^{th}$ column in (b)). The results also demonstrate bi-directional robustness as the change masks are of similar pattern within each source-target pair.

\subsection{Results on the Object Change Detection Dataset}
In addition to the synthetic tabletop dataset, we further evaluate our algorithm on the more challenging real-world object change detection dataset, which offers scene settings with object changes of more diverse sizes and layouts.

Quantitatively, Table~\ref{result} compares the results of our approach in terms of the five metrics against those of the volumetric/point-based approaches Octomap~\cite{langer2017fly} and Meta-room~\cite{ambrucs2014meta}, and the best results of the approach proposed by ~\cite{langer2020robust}.  The results are computed by projecting the ground truth point clouds into SDF voxels and determining the change state of each point according to that of its corresponding voxel indicated by the 2D change mask from our approach. Note that following dataset definition, we manually exclude all detected changed points resulting from moved furniture and decoration from evaluation. 

Moreover, to demonstrate the effectiveness of our blob-level curvature-based SDF description for robust change detection, we provide another baseline (\emph{FPFH} in Table~\ref{result}) with a point-wise variant of the proposed method by replacing  the 3D voxel validation step~\ref{3d}  with the point-based FPFH~\cite{rusu2009fast} feature matching using the Open3D~\cite{zhou2018open3d} implementation. As our selected key voxels are \emph{not} located on object surfaces, where off-the-shelf point feature extractors cannot be directly applied, FPFH features are extracted for every point in the original point cloud that contributes to the fusion of the SDF. A point is marked as changed if its source FPFH feature cannot be matched in its target neighborhood. 

Here, Fig.~\ref{keypts} illustrates the key voxel distribution and the false positive points detected by FPFH matching for two unchanged sub-scenes: a single green object and two bottle standing closely against a wall. In (b), near-surface key voxels (within 1.5 SDF voxel size to an object point, shown in blue) are distributed around the object surface, giving good characterization of the object geometry, while key voxels farther away from the surface are more frequently witnessed in spaces amid surfaces, e.g., the area around the top of the shorter bottle and the left gap between the bottles and the wall, acting to unravel the spatial relations of these adjacent surfaces. The effectiveness of considering both object surfaces and inter-surface regions is then demonstrated by (c). While our method correctly recognizes the two scenes as unchanged, FPFH shows a small ratio of false positive points for the less noisy, single-object scenario but induces considerable amounts of false positives for the two-bottle case given a partial and warped reconstruction of the shorter bottle and the wall. 

From Table.~\ref{result}, we see that our approach achieves the highest values in terms of the five aforementioned metrics in most scenes. The point-wise FPFH matching baseline, while not eligible for wrongly detected clusters measurements as no cluster-level operations are involved, results in worse performance in the rest of the four metrics. This can be ascribed to its sensitivity to reconstruction noises, e.g., residual points or warpings that are prevalent around boundaries.  

In comparison to the baseline approaches,  our better performance could be attributed to the more distinct object contours and more robust neighborhood geometry verification enabled by the PlaneSDF representation. First, finding intersections between the preliminary change mask and the object map ensures that most of voxels extracted for 3D validation belong to \emph{part of} an object and \emph{all} voxels of the potentially changed objects are selected for 3D validation, hence  unaffected by the common artifacts, e.g., noisy and incomplete object boundaries, in 3D clustering and segmentation in~\cite{langer2020robust}. Second, local geometry verification, as opposed to point-wise nearest neighbor searching, offers additional robustness for detecting smaller objects and rejecting false positives, especially in the face of undesired point cloud residuals, such as when reconstruction quality is poor and objects are close to fixed structures such as walls.

Qualitatively, Fig.~\ref{panorama} and Fig.~\ref{compare} display examples of qualitative change detection results of each of the five scenes. From Fig.~\ref{compare}, we can see that the proposed algorithm is able to extract point clouds belonging to most of the newly introduced objects, with some points missing from the planar parts that are attached to the plane, such as the bottom of the skillet in the kitchen scene (the last row of Fig.~\ref{compare}).

While the proposed algorithm has been shown to be effective in object change detection both quantitatively and qualitatively,  we point out the failure case as when the height discrepancy between the object and the plane is ambiguous. Two typical examples within the dataset are: (1) The new object is partially occluded by a fixed structure in the height direction, e.g., the baseball placed under the table is missing from detection as its height is not correctly reflected in the height map. (2) The object is close to some noisy plane boundaries such as  those caused by non-rigid deformation, e.g., missing object detection on the sofa (first row in Fig.~\ref{compare}) and our lower precision scores on the ``small room'' and ``living room'' scenes with new objects on the sofa.

\section{Conclusions}
In this paper, we have presented a new approach for change detection based on the newly proposed  PlaneSDF representation. By making the most of the plane-supporting-object structure, our approach decomposes the common noise-sensitive global scene differencing scheme in a local plane-wise and object-wise manner, demonstrating enhanced robustness to measurement noises and reconstruction errors on both synthetic and real-world datasets.
\bibliographystyle{ieeetr}
\bibliography{ref}  
\end{document}